\pdfoutput=1

\documentclass[11pt]{article}

\usepackage{acl}
\usepackage{times}
\usepackage{latexsym}
\usepackage[T1]{fontenc}
\usepackage[utf8]{inputenc}
\usepackage{microtype}
\usepackage{booktabs}
\usepackage{graphicx}
\usepackage{multirow}
\usepackage{amsmath}

\title{Enhancing the General Agent Capabilities of Low-Parameter LLMs through Tuning and Multi-Branch Reasoning}

\author{Qinhao Zhou\textsuperscript{1}\quad Zihan Zhang\textsuperscript{1}\quad Xiang Xiang\textsuperscript{1}\thanks{$^*$Corresponding author (e-mail: \url{xex@hust.edu.cn}); also with Peng Cheng Laboratory, Shenzhen, China.}\quad \\
\textbf{Ke Wang}\textsuperscript{2}\quad \textbf{Yuchuan Wu}\textsuperscript{2}\quad \textbf{Yongbin Li} \textsuperscript{2}\\
\textsuperscript{1} National Key Lab of MSIIPT, School of Artificial Intelligence and Automation, \\Huazhong University of Science and Technology, Wuhan, China \\
\textsuperscript{2} DAMO Academy, Alibaba Group, Beijing, China \\
}

\begin{document}
\maketitle
\begin{abstract}
 Open-source pre-trained Large Language Models (LLMs) exhibit strong language understanding and generation capabilities, making them highly successful in a variety of tasks. 
 However, when used as agents for dealing with complex problems in the real world, their performance is far inferior to large commercial models such as ChatGPT and GPT-4. 
 As intelligent agents, LLMs need to have the capabilities of task planning, long-term memory, and the ability to leverage external tools to achieve satisfactory performance. 
 Various methods have been proposed to enhance the agent capabilities of LLMs. On the one hand, methods involve constructing agent-specific data and fine-tuning the models. 
 On the other hand, some methods focus on designing prompts that effectively activate the reasoning abilities of the LLMs. We explore both strategies on the 7B and 13B models. 
 We propose a comprehensive method for constructing agent-specific data using GPT-4. 
 Through supervised fine-tuning with constructed data, we find that for these models with a relatively small number of parameters, supervised fine-tuning can significantly reduce hallucination outputs and formatting errors in agent tasks. 
 Furthermore, techniques such as multi-path reasoning and task decomposition can effectively decrease problem complexity and enhance the performance of LLMs as agents. 
 We evaluate our method on five agent tasks of AgentBench and achieve satisfactory results.

\end{abstract}

\section{Introduction}

\begin{figure}[t]
  \centering 
    \includegraphics[width=0.48\textwidth]{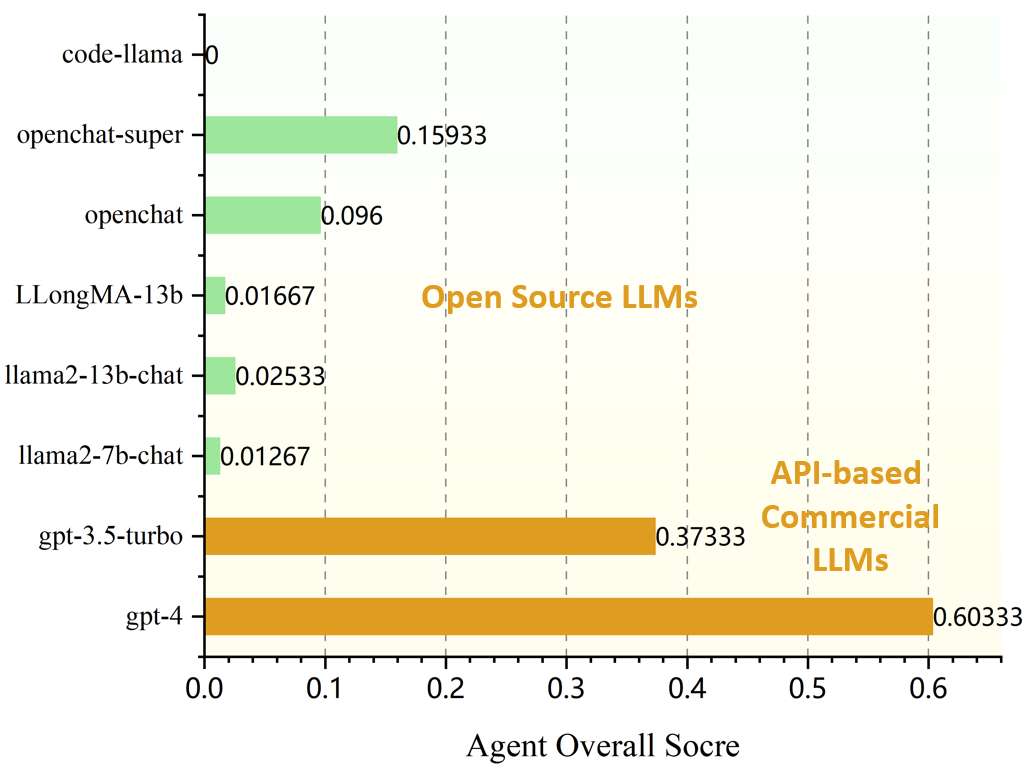}
  \caption{ 
   The agent performance of open-source LLMs and commercial LLMs. Agent Overall Score is the average accuracy of several agent tasks.  
  }
  \label{intro}
\end{figure}
Large Language Models (LLMs) have been extensively employed in a wide range of natural language processing tasks, yielding groundbreaking achievements. 
Furthermore, LLMs have demonstrated their capability to undertake more challenging tasks, such as functioning as AI agents. 
Unlike conventional reasoning tasks, an AI agent is an entity that needs to interact with the human or external environment, draw inferences, and judge subsequent actions based on feedback. 
Each single task typically involves multiple rounds of dialogue to accomplish. 
For instance, in a home environment, an agent may be tasked with various household tasks that require continuous interaction with the environment. The agent needs to evaluate its actions based on the feedback from the environment and make timely adjustments to its strategies.    
Traditional AI agents are usually effective in specific domains or environments, but their generalization and adaptability are obviously insufficient \cite{liu2023agentbench}. 

In recent years, an increasing number of work \cite{brown2020language, openai2023gpt, qin2023toolllm, shinn2023reflexion, zhu2023ghost} have demonstrated that LLMs possess strong capabilities in reasoning, planning, memory, and utilizing external tools. 
This has propelled LLMs towards becoming more generalized and adaptive agents. 
Recently, AgentBench \cite{liu2023agentbench} conducts extensive evaluations of both commercial and open-source LLMs on eight different agent tasks. The results reveal that commercial API models show superior agent capabilities. 
In addition, work such as AutoGPT \cite{gravitas2023auto} and GPT-Engineer \cite{osika2023gpt} also use LLMs as agents to build a complete framework for solving complex real-world problems. 
However, open-source models, especially those with smaller parameter sizes, still have substantial potential for enhancement.
As shown in Fig.~\ref{intro}, the average performance of 7B and 13B LLMs on each agent task is significantly lower than the commercial models. 

Unlike commercial LLMs, small-scale open-source LLMs are relatively inefficient in general knowledge \cite{peters2019knowledge}.
Besides, lower parameter sizes limit reasoning and memory capacity, often leading to hallucinations in the agent dialogue process \cite{zhang2023siren}. However, in practical applications, LLMs with 7B and 13B parameters are the most widely used due to their relative ease of deployment and fine-tuning. Therefore, enhancing the capabilities of such LLMs is of great practical significance. Currently, studies on LLMs agents or enhancing model reasoning capabilities \cite{xi2023rise, wang2023survey} primarily focus on large-scale models. The investigation of agent capabilities on 7B and 13B LLMs is still in its early stages of exploration. 
As explained, a proficient agent requires task-planning abilities, proficiency in utilizing external tools, and long-term memory capabilities. Task planning refers to the ability of the model to decompose large-scale tasks into manageable sub-goals, facilitating efficient handling of complex tasks. 
Long-term memory capabilities reflect the ability of the LLMs to retain and recall historical information during their interactive processes with the environment. Considering these abilities, we propose a method to enhance the performance of 7B and 13B LLMs on agent tasks. 

In our proposed approach, We focus on enhancing the agent capabilities of LLMs from two key aspects. First, improving the agent capabilities through Supervised Fine-Tuning (SFT). This approach fundamentally enhances the LLMs themselves. 
Unlike general reasoning tasks, an agent's role goes beyond planning and reasoning. It also involves continuous interaction with the environment or humans to execute subsequent actions until a desired outcome is achieved. 
To improve the agent abilities of LLMs, it is essential to train them on diverse datasets that reflect the full range of interactive behaviors between the agent and the environment. This involves constructing data that not only records the actions taken by the agent but also captures the internal thought processes and decision-making. Additionally, the environment should provide meaningful feedback to guide the learning of the agent. We propose to use GPT-4 \cite{openai2023gpt} to construct data. By designing a framework that involves GPT-4 engaging the multi-turn dialogues, we can generate conversational data that captures the interaction between different roles. During these conversations, GPT-4 can take on different roles, such as playing the part of an agent, a user, or the environment, and actively participate in dynamic exchanges. In addition, we incorporate a significant amount of general instruction tuning data into the constructed dataset to preserve the general capabilities of the LLMs. 

 Besides, we optimize the reasoning path through task decomposition and backtracking. Inspired by Chain of Thought \cite{wei2022chain}, significant efforts have been dedicated to activating the reasoning ability of the LLMs. For instance, ReAct \cite{yao2022ReAct} integrates the thinking process into the task of multi-step reasoning. ToT \cite{yao2023tree} uses depth-first and breadth-first traversal of reasoning nodes, which is more conducive to finding the optimal solution. We migrate the idea of ToT to the agent tasks and combine it with task decomposition and backtracking. Task decomposition leverages the task planning capability of the LLMs to decompose complex and lengthy tasks into several smaller subtasks. Considering that it is difficult for LLMs to find optimal answers or complete tasks through a single reasoning path, we introduce a judgment process where the reasoning process goes back to the starting point, termed backtracking. Through the integration of task decomposition and backtracking, we aim to enhance LLMs' ability to handle complex tasks effectively.  

The main contributions of this paper are: 1) We explore the capabilities of 7B and 13B open-source LLMs as agents, exploring their potential in performing agent tasks. 2) We propose supervised fine-tuning with specific agent data as a fundamental approach to improving the capability of open-source LLMs as agents. To achieve this, we develop a method for constructing agent data. 3) We find that task decomposition and backtracking are effective approaches for addressing complex agent tasks. We conduct experiments on AgentBench and achieve promising results.

\section{Related Works}
\textbf{Planning and Reasoning.} Planning and reasoning are crucial capacities for agents to solve complex tasks. Through the in-context of the thinking chain, Chain-of-Thought \cite{wei2022chain} activates the reasoning capabilities of LLMs and enables the generation of intermediate thought processes before producing answers. Some other strategies have also been proposed to further enhance the thinking process of models. For example, SC \cite{wang2022self} leverages the self-consistency of LLMs by generating multiple thinking chains and determining the final answer through voting. Reconcile \cite{chen2023reconcile} enhances the reasoning capabilities of LLMs through multiple rounds of discussions and using confidence-weighted voting. Besides, self-polish \cite{xi2023self}, and self-refine \cite{madaan2023self} augment the thinking process of LLMs from other perspectives. Furthermore, ToT \cite{yao2023tree} explores the abstracting reasoning process into deep tree search. In addition, there are some works \cite{zhang2023multimodal} that apply the idea of chain thinking to multi-modal tasks. 

\textbf{Large Language Model as Agent.} With the rapid advancement of LLMs, extensive research has been conducted to explore their powerful capabilities in planning and reasoning \cite{xi2023rise, wang2023survey}. This has opened up the possibility of employing LLMs as agents. On the one hand, there have been several efforts to apply LLMs to various agent tasks and construct agent simulation frameworks. On the other hand, several works \cite{xu2023rewoo, kim2023language}, such as ReAct \cite{yao2022ReAct}, have focused on incorporating reasoning and deliberation into the agent process for LLMs. In addition, some works apply the reasoning methods to the agent interaction process. PET \cite{wu2023plan} applies task decomposition to the household agent environment, which is helpful for LLMs to complete complex tasks. LATS \cite{zhou2023language} and RAP \cite{hao2023reasoning} apply Monte Carlo tree search to the agent reasoning process. It is advantageous to find better answers compared with ToT. In addition, research works such as AutoGPT \cite{gravitas2023auto} and GPT-Engineer \cite{osika2023gpt} utilize commercial LLMs as agent core of their frameworks, enabling the development of comprehensive agent architectures to tackle complex real-world problems.  

\textbf{Instruction Tuning for Language Model.} Instruction tuning plays a crucial role in training LLMs. After pre-training with massive unsupervised data, LLMs acquire a substantial amount of knowledge and process language understanding and generation capabilities. Further supervised instruction fine-tuning \cite{zhang2023instruction, dong2022survey} is conducted to align the model with human instructions and generate outputs that better align with human preferences. Instruction tuning mainly focuses on constructing complex and diverse general-purpose tasks to train LLMs to answer questions in a human manner. For example, FLAN \cite{wei2021finetuned} and T0 \cite{sanh2021multitask} construct a multi-task instruction tuning dataset using massive publicly available datasets. The fine-tuned model shows strong zero-shot generalizability. In addition to utilizing existing datasets, another common approach is to generate data using commercial LLMs. Self-Instruct \cite{wang2022self, peng2023instruction} leverages GPT-4 to generate a large amount of diverse data, given a few seed tasks. These data are used for fine-tuning open-source LLMs and get significant improvements in various tasks. To enhance the agent capability of LLMs, AgentTuning \cite{zeng2023agenttuning} utilizes commercial LLMs to construct data in specific agent environments containing multi-turn dialogues.

\section{Methodology}
In this section, we first give a formal definition of LLMs as agents. Then, we introduce the two components of our approach. In the first part, we construct agent-tuning data to fine-tune LLMs with parameter-efficient tuning methods. This is a way to fundamentally improve the capabilities of LLMs. In the second part, we propose enhancing the reasoning capabilities of LLMs through task decomposition and backtracking. 

\subsection{Problem Formulation}
For a given agent task, the interaction trajectory of LLMs as agents can be represented as a dialogue history $(e_1, a_1,...,e_n, a_n)$. During this process, there are typically two roles involved: environment and agent. $e_i$ represents the hints and feedback from the environment and the agent engages in thinking and actions represented as $a_i$. Each dialogue track corresponds to a final reward $r\in[0, 1]$, which reflects the completion of the task. 

\subsection{Supervised Tuning with General and Constructed Agent Data}
We observe a significant disparity in the agent capabilities between the open-source 7B and 13B LLMs and the commercial models. In the dialogue process, open-source models often exhibit issues such as formatting errors, getting stuck in infinite loops, and generating hallucinatory outputs. To reduce the occurrence of the above issues, a fundamental approach is to fine-tune the LLMs with appropriate data. However, the agent is engaged in multi-turn dialogues and interacts with specific environments, which is different from currently available open-source general-purpose instruction data. To solve this challenge, we leverage commercial models API to construct agent-specific data and merge them with general instruction datasets to fine-tune the low-parameter LLMs. 

\begin{figure}[t]
  \centering 
    \includegraphics[width=0.49\textwidth]{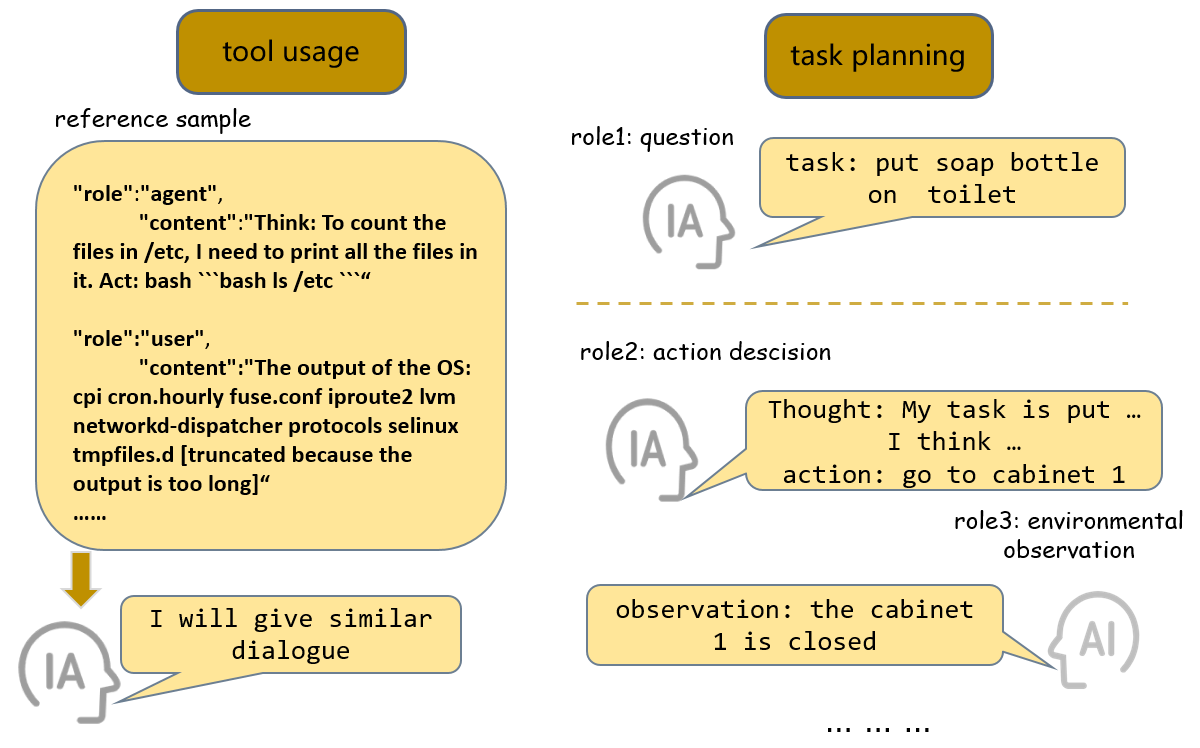}
  \caption{ 
   The process of constructing agent data. For task planning and external tool usage capabilities, we use two strategies, respectively. 
  }
  \label{agent-data-sft}
\end{figure}

 As agents, LLMs need to possess three fundamental capabilities: task planning, long-term memory, and tool usage. To enhance the task planning capabilities of LLMs, we take ALFworld \cite{shridhar2020alfworld} as an example to construct data with interactive trajectories. Unlike current methods of constructing data using models like GPT-3.5 \cite{gpt-3.5}, data for agents should not only involve multi-turn dialogues but also need to reflect task planning and trajectory. Therefore, we meticulously design the construction process of the dataset, dividing the process of each piece of data into three steps. It includes task construction, trajectory interaction, and manual filtering. This approach ensures that each piece of data captures the necessary elements for training agents effectively.
We utilize GPT-3.5 or GPT-4 to generate questions and interaction trajectories and this process can be easily extended to other agent tasks. As illustrated in Fig.~\ref{agent-data-sft} right, to generate a complete interaction trajectory, we simulate GPT playing three distinct roles in a household environment. These roles are named as \textit{question generator}, \textit{action maker}, and \textit{environmental agent}. 

First, we randomly initialize a specific room environment, determining the number and placement of household items. The \textit{question generator} role is then responsible for generating intelligent household-related questions based on the provided environment. Subsequently, the \textit{action maker} role continuously offers its thoughts and actions based on the environment feedback, simultaneously, the \textit{environment agent} role provides reasonable feedback and cues corresponding to the actions taken in each step. These two roles continue to interact until the problem is completed or the maximum number of interactions is reached, thus generating a complete trajectory. However, as there is no assurance of the logical consistency of the \textit{environment agent}'s feedback and the \textit{action maker}'s actions, manual screening is required after the data is generated.  

In addition to agent tasks that focus on task planning, there are also agent tasks such as Operating System, and WebShop \cite{yao2022webshop} that have fewer dialogue rounds and prioritize the use of external tools. For this type of task, we draw on the idea of in-context learning. Specifically, as shown in Fig.~\ref{agent-data-sft} left, we provide GPT with examples with complete reasoning trajectories to enable it to imitate. Subsequently, we manually filter and select logically consistent data from generated outputs. We expect to use this type of data to improve the retrieval capabilities and tool usage capabilities of LLMs.

Existing work on agent fine-tuning \cite{zeng2023agenttuning} shows that using only agent data to fine-tune LLMs compromises their generalizability. Therefore, we mix some general instruction tunning data into our agent data when fine-tuning LLMs. Suppose $M_{\theta}$ represents pre-trained LLMs and the $M_{\theta}(y|x)$ represents the probability distribution of output $y$ when given history $x$. We consider two datasets: the agent data $D_{agent}$ and the general instruction tuning data $D_{general}$. We optimize the loss function as follows:

\begin{equation}
\begin{split}
    \mathcal{L}(\theta) &= \lambda \cdot  \mathbf{E}_{(x, y)~D_{agent}}[log M_{\theta}(y|x)] \\
    & + (1 -  \lambda) \cdot \mathbf{E}_{(x, y)~D_{general}}[log M_{\theta}(y|x)].
\end{split}
\end{equation}

Where $\lambda \in [0, 1]$ denotes the mix ratio of the two datasets. A larger $\lambda$ means that the LLMs are inclined to specific agent capabilities, whereas a small $\lambda$ makes LLMs more inclined to general capabilities. We observe that deterioration of the general ability of LLMs will also decrease the agent ability, so we set a small value for $\lambda$. This is identical to AgentTuning \cite{zeng2023agenttuning}. In the experimental section, we analyze different values of $\lambda$.

In the context of fine-tuning strategiy, we adopt Low-Rank Adaptation (LORA) \cite{hu2021lora} fine-tuning which is based on making low-rank modifications to the weight matrices in LLMs. For each linear layer in the model, the original weight matrix \( W \) is adjusted to \( W + \Delta W \), where \( \Delta W \) is generated through the product of low-rank matrices as $\Delta W = A \times B$, where \( A \) and \( B \) are low-rank matrices, with ranks significantly smaller than the rank of the original weight matrix \( W \). 



\subsection{Multi-Path Reasoning under Task Decomposition}


Recently, because it is difficult for a single agent to complete complex multi-step tasks, more and more work tends to involve multi-agent collaboration, allowing models to play different roles to jointly advance tasks \cite{qiao2024autoact}. We take a similar approach. On the one hand, we we instruct LLMs to generate multiple available actions in each reasoning step. On the other hand, we employ a judge model to select one action from the provided set and continue the reasoning process until a final output is obtained. 

\begin{figure}[t]
  \centering 
    \includegraphics[width=0.48\textwidth]{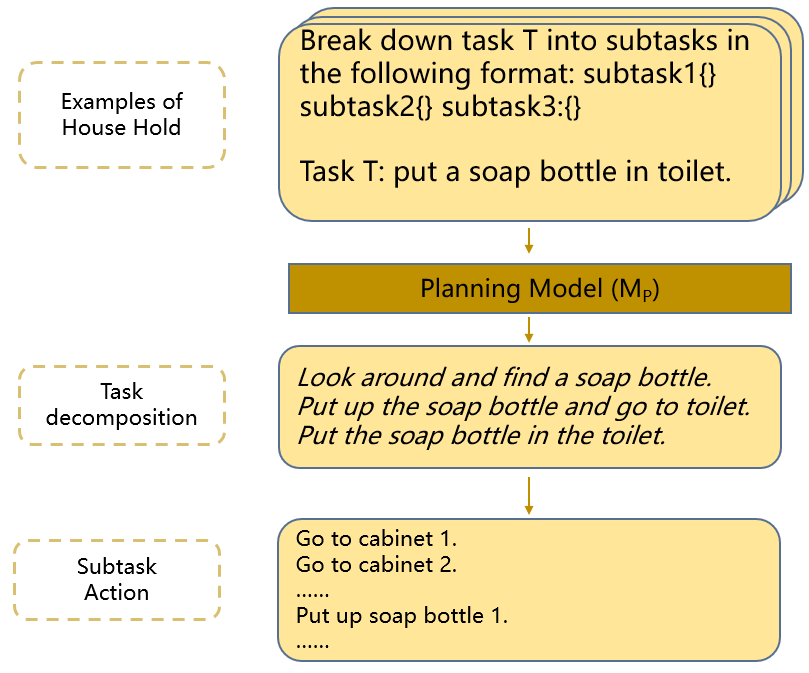}
  \caption{ 
   The process of task decomposition. The planning model breaks the entire task into several small subtasks.
  }
  \label{task_decomposition}
  \vspace{-3mm}
\end{figure}

For LLMs with small parameter sizes, due to their limited long-term memory capacity, it is challenging for them to handle complex long dialogue tasks. To address this issue, we employ a task decomposition strategy, where complex tasks that require multiple steps are broken down into simpler subtasks. We use another LLM with the same number of parameters as our planning module and we name it as $M_{p}$. For a given task $\mathcal{T}$, we compose query prompt $P_{sub}$ as "break down the task $\mathcal{T}$ into subtasks in the following format...". The $M_{p}$ will generate a sub-task list $S_{\mathcal{T}}= \{s_1,...,s_k\}$. $k$ is the number of sub-tasks and to avoid an excessive number of subtasks, we typically set $k$ to $3$. For example, for task $\mathcal{T} = $"put a soap bottle in the toilet", the LLMs can describe three steps as $s_1 = $ "look around and find a soap bottle", $s_2 = $ "take up the soap bottle and go to the toilet", $s_3 = $ "put the soap bottle in the toilet". Then, the agent will complete it one by one according to the subtask list $S_{\mathcal{T}}$. We introduce another LLM as judgment module $M_{jdg}$ to judge the completion of each subtask. For subtask $s_t$, we compose the judge prompt $P_{jdg}$ as "Judge whether the subtask is completed, output Yes or No", each time the agent executes a step, we feed $P_{jdg}$ to a LLM and get the output of "Yes" or "No" until the subtask is completed. 

Agent tasks in the real world are often complex and one single reasoning path may not yield the optimal answer. Inspired by the reflective ability in human thinking processes, we propose to take multi-path reasoning with LLMs. We call this method \textit{backtracking}. When a particular reasoning path yields a suboptimal output, we compose a backtracking prompt as "it was observed that the answer was not the optimal choice for task $\mathcal{T}$...". We also prompt the LLMs to eschew reasoning paths that have been previously deduced. To this end, we compose the prompt as "it is important to note that actions should be adjusted appropriately based on the historical information" and we splice this prompt behind the backtracking prompt. Furthermore, backtracking and task decomposition are not mutually exclusive and can be applied together in the reasoning process of LLMs. We find that task decomposition is more effective for agent tasks that emphasize planning abilities, while backtracking is more effective for agent tasks that emphasize API invocation capabilities. 

\begin{figure}[t!]
  \centering 
    \includegraphics[width=0.48\textwidth]{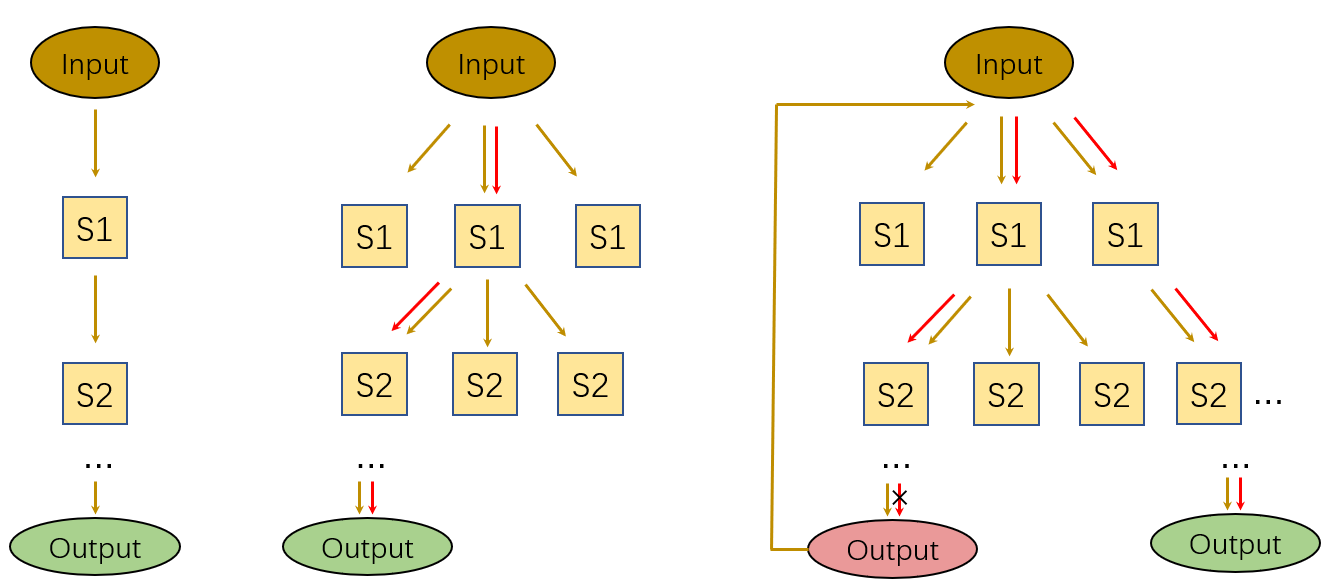}
  \caption{ 
   The comparison of different reasoning methods. From the left to right are Input Output (IO), ToT and our method.
  }
  \label{ablation_data_num}
\end{figure}

Overall, our method is divided into two parts. The first part uses commercial LLMs to construct agent data and employs SFT to fundamentally enhance the agent capabilities of low-parameter LLMs. In the second part, while keeping the LLMs unchanged, it maximizes the activation of the agent capabilities by incorporating multi-path reasoning and task decomposition. For 7B and 13B LLMs, common issues such as hallucinatory outputs and forgetting errors often occur. By fine-tuning the LLMs on domain-specific data that adheres to the desired format, these issues can be significantly mitigated. For reasoning problems with vast search spaces, finding the optimal solution through a single inference path is challenging. This issue cannot be effectively addressed through supervised fine-tuning alone. However, by introducing techniques such as multi-path reasoning and task decomposition, the complexity of the problem can be reduced, facilitating the identification of the optimal solution. 

\setlength{\tabcolsep}{3pt}
\renewcommand\arraystretch{1.0}
\begin{table*}[t]
\centering
\scalebox{0.96}{
\begin{tabular}{cccccccc} \toprule
               &            Data type                   & Operating System    & DataBase                   & Webshop   & ALFWorld & Mind2web                   & Avg. $\uparrow$      \\ \hline \hline
GPT-4          &                               & 42.4  & 32                   & 61.1 & 78 & 29                   & 48.50     \\
GPT-3.5-turbo  &                               & 32.6  & 36.7                 & 64.1 & 16 & 20                   & 33.88    \\
claude         &                               & 9.7   & 22                   & 55.7 & 58 & 25                   & 34.08    \\ \hline
llama2-chat w/o sft       &                               & 3.8   & 2.66                 & 0    & 0  & 5.68                 & 2.43    \\
codegen-struct & \multirow{2}{*}{code}         & 3.8   & 1.3 & 0    & 0  & 0                    & 1.27 \\

alpaca-code    &                               & 3.8   & 1.3                  & 4.20  & 0  & 5.68                 & 2.99    \\

open-assistant & dialog                        & 0     & 2.67  & 2.70  & 0  & 3.41 & 1.76      \\
alpaca         & \multirow{3}{*}{instro+agent} & 15.38 & 3.33                 & 31.10 & 0  & 8.52                 & 11.67   \\
agenttuning    &                               & 15.38 & 38.30                 & 32.60 & 10 & 7.38                 & 20.73   \\
ours           &                               & 11.54   & 27.0                 & 34.53  & 10  & 9.66                  & 18.33    \\
\bottomrule
\end{tabular}
}
\caption{The experimental results of fine-tuning LLMs with different instruction tuning datasets on AgentBench tasks. We use llama2-7b-chat as the base model.}
\label{sft}
\vspace{-3mm}
\end{table*}

\section{Experiments}
\textbf{Agent Datasets:} We select five tasks from AgentBench benchmark \cite{liu2023agentbench}: ALFWorld, WebShop, Mind2Web, Operating System, and Database. Next, we will introduce each agent task one by one in detail.  

\textbf{ALFWorld} is designed to evaluate the planning ability of LLMs in a simulated home environment. The model needs to make decisions and execute actions through a text interface based on the environment description and target instructions, and dynamically adjust the plan to complete the task. 

\textbf{WebShop} aims to evaluate the performance of LLMs in a simulated online shopping environment that mimics a real e-commerce website.The goal of the evaluation is to require LLMs to shop in a virtual shopping environment according to instructions and select products that meet desired attributes.


\textbf{Mind2Web} is a general web agent evaluation benchmark designed to evaluate the ability of LLMs to perform complex tasks on websites in different domains. The dataset covers a cross-domain test set across multiple websites. Each task includes a task description, a reference action sequence, and web page information and is designed to test the performance of LLMs in web browsing and interactive environments. 

\textbf{Operating System} is designed to evaluate the ability of LLMs to perform tasks in the Bash environment of a real operating system. Tasks includes question answering and action, where the model needs to generate commands to solve a problem or perform an action. 

\textbf{DataBase} is designed to evaluate the ability of LLMs to operate via SQL on real databases. The dataset contains a diverse set of instructions and databases, created by combining multiple existing datasets and performing data augmentation. 

\textbf{Implementation details:} We use AgentBench as our benchmark and conduct experiments based on it. For 13B models, we choose OpenChat. OpenChat is a series of open-source LLMs fine-tuned on diverse and high-quality datasets of multi-round conversations. We select two models, openchat-v3.2 and openchat-v3.2-super for experiments. For the 7B models, we select llama2 and agentlm \cite{zeng2023agenttuning} for experiments. We use the fastchat framework to deploy LLMs and we use four RTX 4090 NVIDIA GPUs. See also the project page\footnote{\url{https://github.com/HAIV-Lab/LLM-TMBR}}.

\subsection{Experimental Results}
\textbf{Supervised fine-tuning with constructed dataset.} The experiments of supervised fine-tuning are shown in Tab.~\ref{sft}. We fine-tune the 7B model on various instruction-tuning datasets and test it on five agent tasks. It can be seen that fine-tuning on various instruction datasets has a positive effect on improving the capabilities of agents. Among them, we find that fine-tuning the LLMs using code-type instructions has shown relatively limited effectiveness in improving agent capabilities. For example, after fine-tuning on alpaca-code dataset, the performance of llama2 on operating system task does not improve, and its performance on database tasks actually declined by $1.33\%$. We analyze that although code-type data can enhance the understanding of the code of LLMs, it lacks dialogue processes and the decomposition of complex problems. Similar to code-type data, fine-tuning LLMs on regular dialog data alone is not an appropriate choice for enhancing its agent capabilities. For instance, after fine-tuning on Open-Assistant, llama2 exhibited a decrease in performance on operating system task and a lower improvement on the webshop task compared to other datasets. 


\setlength{\tabcolsep}{4pt}
\renewcommand\arraystretch{0.90}
\begin{table*}[t]
\centering
\begin{tabular}{ccccccl} \toprule
\multicolumn{1}{l}{}  Size &   LLMs   &Methods  &
\multicolumn{1}{l}{}  Webshop & ALFWorld & Operate System & \multicolumn{1}{c}{Avg. $\uparrow$} \\
\hline \hline
\multirow{8}{*}{13B} & \multirow{4}{*}{openchat\_v3.2}        & IO                   & 1        & 0          & 0              & 0.33                \\
                                 &                                        & CoT                  & 19       & 0          & 0              & 6.33                \\
                                 &                                        & ReAct                & 26       & 5          & 7.6            & 12.86                \\
                                 &                                        & Ours                 & 27       & 10         & 7.6            & 14.86                \\
                                 \cline{3-7}
                                 & \multirow{4}{*}{openchat\_v3.2\_super} & IO                   & 5        & 0          & 0              & 1.66                \\
                                 &                                        & CoT                  & 23       & 0          & 0              & 7.66                \\
                                 &                                        & ReAct                & 30       & 5          & 3.8            & 12.93                \\
                                 &                                        & Ours                 & 31       & 11         & 3.8            & 15.26                \\
                                 \hline
\multirow{8}{*}{7B}  & \multirow{4}{*}{AgentLM-7B}            & IO                   & 50       & 5          & 3.8            & 20.86                \\
                                 &                                        & CoT                  & 34       & 5          & 7.6               & 19.50                    \\
                                 &                                        & ReAct                & 33       & 0          & 7.6            & 13.53                \\
                                 &                                        & Ours                 & 51       & 0          & 7.6              & 19.53                    \\
                                 \cline{3-7}
                                 & \multirow{4}{*}{llama2-7B}             & IO                   & 0        & 0          & 0               & 0                       \\
                                 &                                        & CoT                  & 4        & 0          & 0               & 1.33                       \\
                                 &                                        & ReAct                & 13.35    & 0          & 7.6            & 6.98                \\
                                 &                                        & Ours                 & 13.40     & 0          & 7.6            & 7.00                       \\
\bottomrule
\end{tabular}
\caption{Experimental results of different reasoning methods on three agent benchmarks.}
\label{reasoning}
\end{table*}

Besides, we find that fine-tuning LLMs on high-quality general instruction tuning datasets can significantly improve its agent capabilities. For example, after fine-tuning with alpaca instruction tuning data, llama2 exhibit significant improvements across multiple agent tasks. In the operating system tasks and webshop tasks, llama2 tuning with alpaca data achieves nearly comparable results to those obtained through agenttuning. Agenttuning is the most effective tuning dataset. It combines GPT-4 assisted trajectory-labeled agent data with general instruction tuning data, resulting in significant improvements for llama2 across different agent tasks. Its performance in the database even exceeds that of the commercial model. 
Fine-tuning the model using our constructed data can also improve the performance of LLMs on agent tasks. Although we construct limited and easy-to-collect data, the performance of LLMs fine-tuned with our data exceeds other datasets on some agent tasks. For example, on operating system tasks, our results are $7.74\%$ higher than code-type datasets and $11.54\%$ higher than dialog-type datasets. Compared with agenttuning, our results are still far behind, which can be attributed to the limited amount of data. In addition, there are fewer complex tasks involving long conversations in our data, which is also one of the reasons. 

\textbf{Reasoning with task decomposition and backtracking.} We compare different reasoning methods on 7B and 13B LLMs, and the results are shown in Tab.~\ref{reasoning}. The 7B LLMs we evaluated are fine-tuned with agent data. AgentLM is fine-tuned with agenttuning data, and llama2 is fine-tuned with the data we constructed. We mainly conduct evaluations on webshop, household and operating system tasks. It can be seen that applying ReAct to various tasks is usually better than direct input and output (IO). For example, on the openchat-v3.2 model, ReAct is $18\%$ higher than IO on webshop. Besides, our method can further achieve small improvements based on ReAct. On the webshop task, our results are on average about $1\%$ higher than the second-best result. And on the household task, our method achieve improvements of $5\%$ and $6\%$, respectively, on the 13B LLMs. 

\begin{figure}[h]
  \centering 
    \includegraphics[height=0.48\textwidth, width=0.48\textwidth] {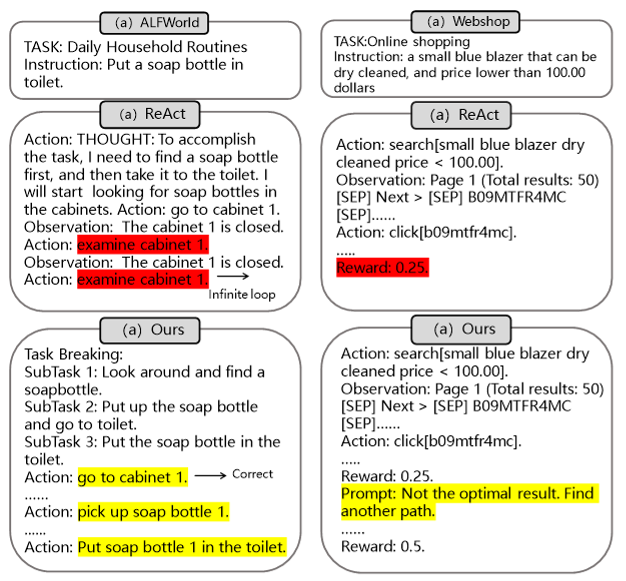}
    \vspace{-3mm}
  \caption{ 
   Comparison of ReAct and our method in agent task reasoning. We show the action and observation in webshop and household tasks.  
  }
  \label{agent_analysis}
  \vspace{-3mm}
\end{figure}

To delve into the impact of different reasoning methods on the results, we compare ReAct and our reasoning process as shown in Fig.~\ref{agent_analysis}. It can be seen that ReAct can prompt LLMs to think in each reasoning step, the models can still experience issues such as getting stuck in infinite loops and suffering from memory confusion. In contrast, on household tasks, since we break down complex tasks into several smaller tasks, model thinking is less error-prone than ReAct. 

\subsection{Ablation Study}
\textbf{The experiments of num path and branch.} "num path" refers to the number of backtracking iterations conducted, with a higher value indicating an increase in the number of reasoning paths explored.  We conduct experiments of "num path" shown in Tab.~\ref{path_branch} left. It can be seen that appropriately increasing "num path" can improve performance, but when "num path" is greater than $2$, performance decreases.
We also conduct the experiments of  "num branch" shown in Tab.~\ref{path_branch} right.
"num branch" is the number of nodes expanded at each reasoning step. It is shown that properly increasing "num branch" can also improve performance: when "num branch" is greater than $2$, performance decreases.

We conduct experiments on the mixing ratio of different general data and agent data as shown in Fig.\ref{lambda}. We find that too much agent data will not bring huge improvements, and general data is equally important.

\begin{table}[t]
\begin{tabular}{cccc} \toprule
num path & Webshop & num branch & Webshop \\ \hline \hline
1        & 20.29   & 1          & 26.00      \\
2        & \textbf{27.00}   & 2          & \textbf{27.00}      \\
3        & 17.84   & 3          & 6.80     \\
4        & 16.67   & 4          & 15.80    \\
\bottomrule
\end{tabular}
\caption{The experimental results of the effect of num path and num branch in our reasoning method.}
\label{path_branch}
\end{table}

\begin{table}[t]
\centering
\begin{tabular}{ccccc} \toprule
 $\lambda$ & Alfworld & Webshop & Mind2web & OS \\ \hline \hline
0.1 & 0.0 & \textbf{38.13} & 6.81 & 0 \\
0.3 & 0.0 & 30.06 & \textbf{7.95} & 0 \\
0.5 & 0.0 & 36.42 & \textbf{7.95} & \textbf{3.8} \\
0.8 & \textbf{5} & 23.35 & 3.97 & 0 \\
\bottomrule
\end{tabular}
\caption{Experimental results after mixing different general data and agent data.}
\vspace{-5mm}
\label{lambda}
\end{table}

\section{Conclusion}
LLMs as intelligent agents have demonstrated powerful agent capabilities. In this work, we explore the 7B and 13B LLMs as agents, and propose to enhance the agent performance of these open-source models by supervised fine-tuning through agent data as well as multi-branch reasoning. SFT can effectively reduce format errors and hallucination output of the LLMs, which not only improves the agent performance but also facilitates the application of various reasoning methods to agent tasks.

\section{Limitations}

This study presents several limitations. First, our experiments are limited to 7B and 13B LLMs, and thus, the applicability of our findings to models of different sizes is not verified. The methods we propose may also not be feasible for all researchers due to the computational demands of fine-tuning larger models. Additionally, measuring reductions in hallucinations and formatting errors is inherently subjective, and the performance metrics used may not fully capture the agent capabilities in complex real-world tasks.

The constructed data for SFT could introduce biases and the potential for model overfitting, limiting the performance of LLMs on unencountered tasks. Moreover, while we implement multi-path reasoning and task decomposition, the strategies for optimizing these techniques are not definitive. Our evaluation on a limited set of tasks does not account for the full range of an agent capabilities, necessitating broader evaluations in future research.

\section*{Acknowledgement}
The funding for this research was generously provided by the Alibaba Innovation Research program under Grant Contract \# CRAQ7WHZ11220001-20978282, Natural Science Fund of Hubei Province (Grant \# 2022CFB823), and HUST Independent Innovation Research Fund (Grant \# 2021XXJS096).

\bibliography{anthology,custom}
\bibliographystyle{acl_natbib}

\end{document}